%% file: IEEE-conference-template-062824.tex
\documentclass[conference]{IEEEtran}
\IEEEoverridecommandlockouts
\usepackage{subfigure}
\usepackage{graphicx}
\usepackage{float} 
\usepackage{wrapfig}
\usepackage{amsmath}
\usepackage{siunitx}
\usepackage{booktabs}
\usepackage{color}
\usepackage{cite}
\usepackage{amsmath,amssymb,amsfonts}
\usepackage{algorithmic}
\usepackage{graphicx}
\usepackage{textcomp}
\usepackage{xcolor}
\usepackage{booktabs} 
\usepackage{colortbl}
\usepackage{tcolorbox}
\def\BibTeX{{\rm B\kern-.05em{\sc i\kern-.025em b}\kern-.08em
    T\kern-.1667em\lower.7ex\hbox{E}\kern-.125emX}}
\begin{document}

\title{Phi-Former: A Pairwise Hierarchical Approach for Compound-Protein Interactions Prediction\\
}

\author{
\IEEEauthorblockN{Zhe Wang\IEEEauthorrefmark{1}\IEEEauthorrefmark{5}, 
Zijing Liu\IEEEauthorrefmark{2}, 
Chencheng Xu\IEEEauthorrefmark{3}, 
Yuan Yao\IEEEauthorrefmark{4}\IEEEauthorrefmark{5}}\\
\vspace{0.3em}
\IEEEauthorblockA{\IEEEauthorrefmark{1}\IEEEauthorrefmark{4}Hong Kong University of Science and Technology, Hong Kong, China\\
\IEEEauthorrefmark{2}International Digital Economy Academy (IDEA), Shenzhen, China\\
\IEEEauthorrefmark{3}Princeton University, Princeton, NJ, USA\\
\IEEEauthorrefmark{5}Corresponding authors.}\\
\vspace{0.2em}
\textit{Emails:} zwangec@connect.ust.hk, liuzijing@idea.edu.cn, cx2502@princeton.edu, yuany@ust.hk
}

\maketitle

\begin{abstract}
Drug discovery remains time-consuming, labor-intensive, and expensive, often requiring years and substantial investment per drug candidate. Predicting compound-protein interactions (CPIs) is a critical component in this process, enabling the identification of molecular interactions between drug candidates and target proteins. Recent deep learning methods have successfully modeled CPIs at the atomic level, achieving improved efficiency and accuracy over traditional energy-based approaches. However, these models do not always align with chemical realities, as molecular fragments (motifs or functional groups) typically serve as the primary units of biological recognition and binding. In this paper, we propose Phi-former, a pairwise hierarchical interaction representation learning method that addresses this gap by incorporating the biological role of motifs in CPIs. Phi-former represents compounds and proteins hierarchically and employs a pairwise pre-training framework to model interactions systematically across atom-atom, motif-motif, and atom-motif levels, reflecting how biological systems recognize molecular partners. We design intra-level and inter-level learning pipelines that make different interaction levels mutually beneficial. Experimental results demonstrate that Phi-former achieves superior performance on CPI-related tasks. A case study shows that our method accurately identifies specific atoms or motifs activated in CPIs, providing interpretable model explanations. These insights may guide rational drug design and support precision medicine applications.
\end{abstract}

\begin{IEEEkeywords}
compound-protein interaction, molecular docking, hierarchical representation learning, functional groups, drug discovery, graph transformer
\end{IEEEkeywords}

\input{sections/introduction}
\input{sections/related}

\input{sections/method}
\input{sections/experiment}

\section{conclusion and future work}

This study demonstrates the importance of modeling motif-level interactions in compound-protein interaction (CPI) prediction for AI-aided drug design. Our Phi-former method employs pairwise hierarchical representation learning to capture atom-atom, motif-motif, and atom-motif interactions through a pre-training framework with two intra-losses and one inter-loss. This enables systematic and mutually beneficial learning across interaction levels. Experimental results show that Phi-former achieves superior performance on CPI tasks, and case studies validate its ability to identify chemically meaningful atoms and motifs activated in interactions. This work highlights the potential of hierarchical deep learning approaches in drug discovery and related fields. While our current framework focuses on CPI tasks, the hierarchical interaction modeling approach is generalizable to other molecular interaction problems.

Future work will expand Phi-former to larger datasets and extend its applicability beyond CPI to drug-drug interaction (DDI) and protein-protein interaction (PPI) tasks, enabling a unified framework for diverse molecular interaction prediction.

\section*{Acknowledgements}
Z.W. and Y.Y. gratefully acknowledge the NSFC/RGC Joint Research Scheme Grant N\_HKUST635/20.






\vspace{12pt}
\end{document}

%% file: sections/introduction.tex
\section{Introduction}

Learning compound-protein interaction (CPI)~\cite{cpi1, cpi2} is essential in drug discovery, involving the identification and characterization of molecular interactions between small organic molecules and target proteins in biological processes~\cite{cpi3, cpi4}. However, experimental characterization remains time-consuming, labor-intensive, and expensive~\cite{b2, cpi5}, necessitating efficient computational methods for CPI-related tasks.

Recent deep learning methods~\cite{b3,b4,b5,b6} represent compound-protein pairs as graph-structured data, modeling atoms as nodes and physical relationships as edges to capture 3D spatial interactions. However, these atom-level approaches overlook that molecular 
motifs serve as the primary units of biological recognition in CPIs. Functional groups determine molecular recognition and binding specificity, dictating interaction types such as hydrogen bonding and hydrophobic interactions. This motif-centric view aligns with chemical intuition and provides interpretable insights for domain experts. Without modeling motif-activated interactions, atom-level predictions may exclude complete functional groups, leading to biased results contradicting biochemical principles.

\begin{figure*}[!htbp]
\centering
\includegraphics[width=\textwidth]{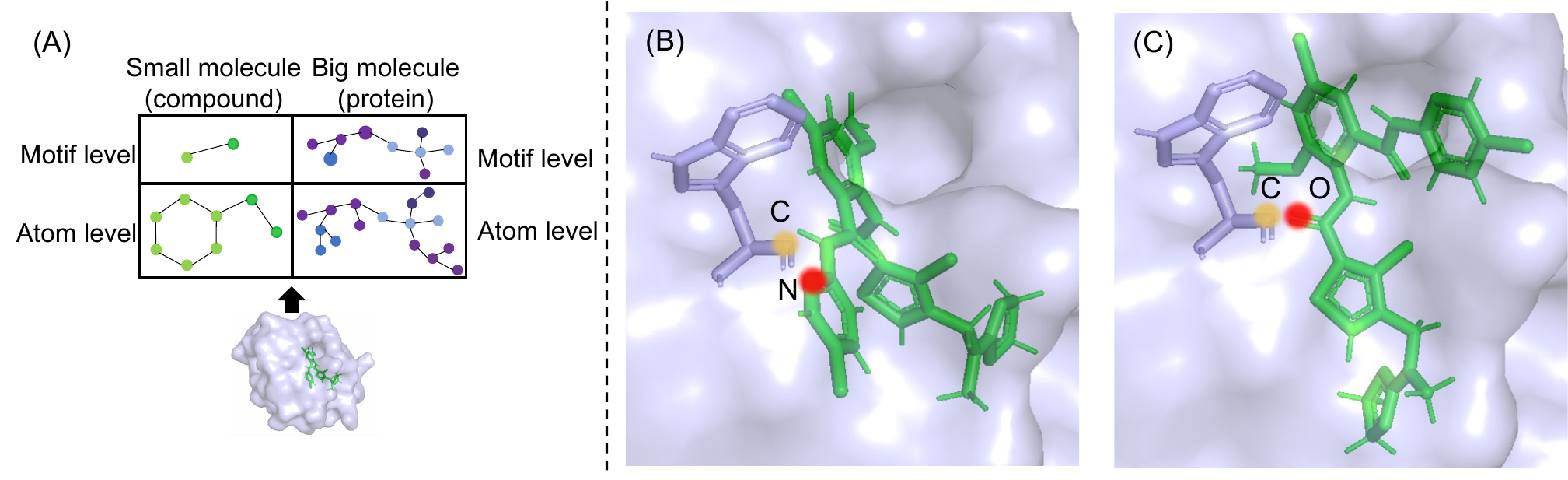}
\caption{(A) The hierarchical levels of protein and compound. The purple surface represents the protein, while the green ligand denotes the compound. (B) The purple ball-and-stick model illustrates a residue within the protein, showcasing the weak interaction between yellow C and red N atoms guided by the affinity between carbonyl and pyridine functional groups. (C) An erroneous binding result predicting weak interaction between the yellow C and red O atoms when functional group constraints are ignored.}
\label{compare}
\end{figure*}

For instance, in Figure \ref{compare}, our objective is to predict the binding conformation of the protein (purple) and the compound (green). Figure \ref{compare}C represents the predicted conformation, while Figure \ref{compare}B serves as the ground truth. Examining this interaction at the atomic level and considering fundamental chemistry principles, hydrogen bonds typically do not prefer forming between carbon (C) and nitrogen (N) atoms. Consequently, neglecting the consideration of molecular motifs may result in an inaccurate predicted binding pose, as demonstrated in Figure \ref{compare}C, where a hydrogen bond is erroneously formed between the C and oxygen (O) atoms. As evident from the ground truth, the interaction between these atoms is not driven solely by individual atomic propensities. Instead, it is facilitated by the affinity of the carbonyl and pyridine functional groups to establish a hydrogen bond between the C and N atoms.

While hierarchical structure modeling exists for single biomolecules \cite{b15,b16}, these approaches are unsuitable for CPI tasks as they lack inter-level constraints. Without constraints from functional groups (Figure \ref{compare}B), erroneous atom-level interactions emerge.

To address these challenges, we propose \textbf{Phi-former}, a \textbf{p}airwise \textbf{h}ierarchical \textbf{i}nteraction representation learning framework employing pre-training and fine-tuning. Phi-former models interactions at both atomic and motif levels, with inter- and intra-losses enabling mutually beneficial learning across hierarchical levels. The model ensures chemical consistency by conditioning atom-atom interactions on motif-motif interactions.

We evaluate Phi-former on binding affinity prediction \cite{b20}, drug-target interaction (DTI) \cite{b19}, and docking pose generation, demonstrating strong performance. Case studies validate our model's ability to capture interactions at atom-atom, motif-motif, and atom-motif levels.

Our contributions include:
\begin{itemize}
\item [-] A hierarchical graph pre-training framework for compound-protein interaction modeling.
\item [-] Inter- and intra-level losses facilitating mutually beneficial learning across interaction hierarchies.
\item [-] Comprehensive evaluation and case studies demonstrating effectiveness and chemical consistency.
\end{itemize}

%% file: sections/related.tex
\section{Related Works}

\subsection{Graph Encoder}
Biomolecules naturally possess graph-based topological structures, prompting extensive use of graph models for molecular representation. Common encoders include Graph Neural Networks (GNNs) such as GCN \cite{b7}, GraphSAGE \cite{b8}, GIN \cite{b10} and MPNN \cite{b9}, as well as graph transformers like Graphormer \cite{b11}, Transformer-M \cite{b12}, and Unimol \cite{b13}, which excel at molecular property prediction, binding affinity prediction, and related tasks.

\subsection{Hierarchical Modeling on Biomolecules}
While most models focus on atomic-level graphs, recent work \cite{b15,b16} demonstrates that incorporating secondary structures enhances performance on biomolecular tasks. This hierarchical molecular graph modeling approach has been extended to CPI tasks \cite{b17}, though these methods primarily focus on single-molecule hierarchies without explicitly modeling pairwise interactions.

\subsection{Binding Affinity Prediction}
Binding affinity quantifies the strength of compound-protein interactions. Sequence-based models like MONN \cite{b22} and TankBind \cite{b4} predict affinity from SMILES \cite{b21} representations, while structure-based methods such as OnionNet \cite{b25}, IGN \cite{b23}, SS-GNN \cite{b5}, and Transformer-M \cite{b12} leverage 3D binding conformations for enhanced modeling. Transformer-M, operating on individual molecular graphs with molecular property pre-training, achieves strong performance on binding affinity tasks.

%% file: sections/method.tex
\section{method}
\begin{figure*}[htbp] 
\centering
\subfigure[pre-training stage]{
\includegraphics[width=0.45\textwidth]{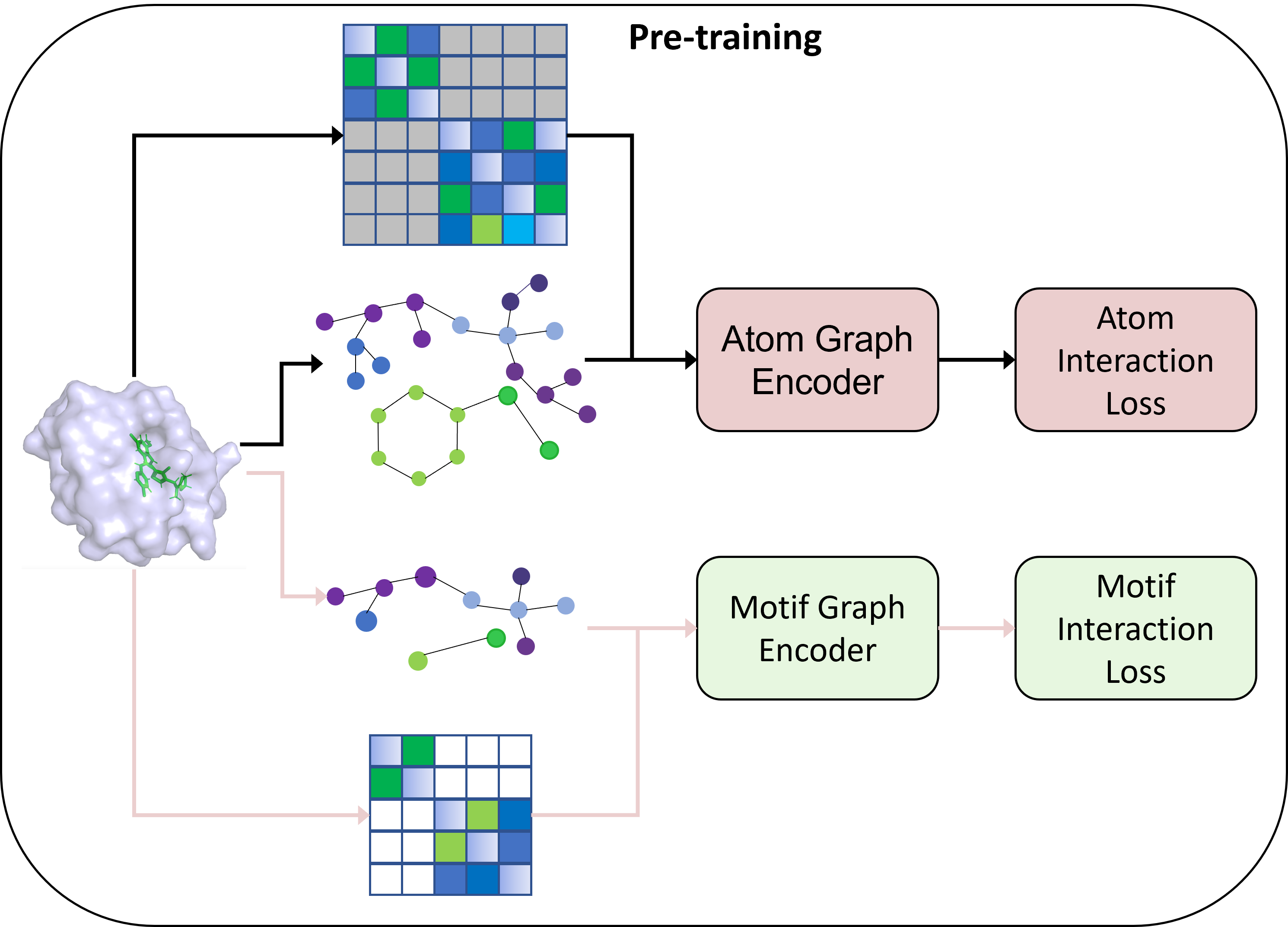}
\label{architecture-1}

}
\subfigure[fine-tuning stage]{
\includegraphics[width=0.5\textwidth]{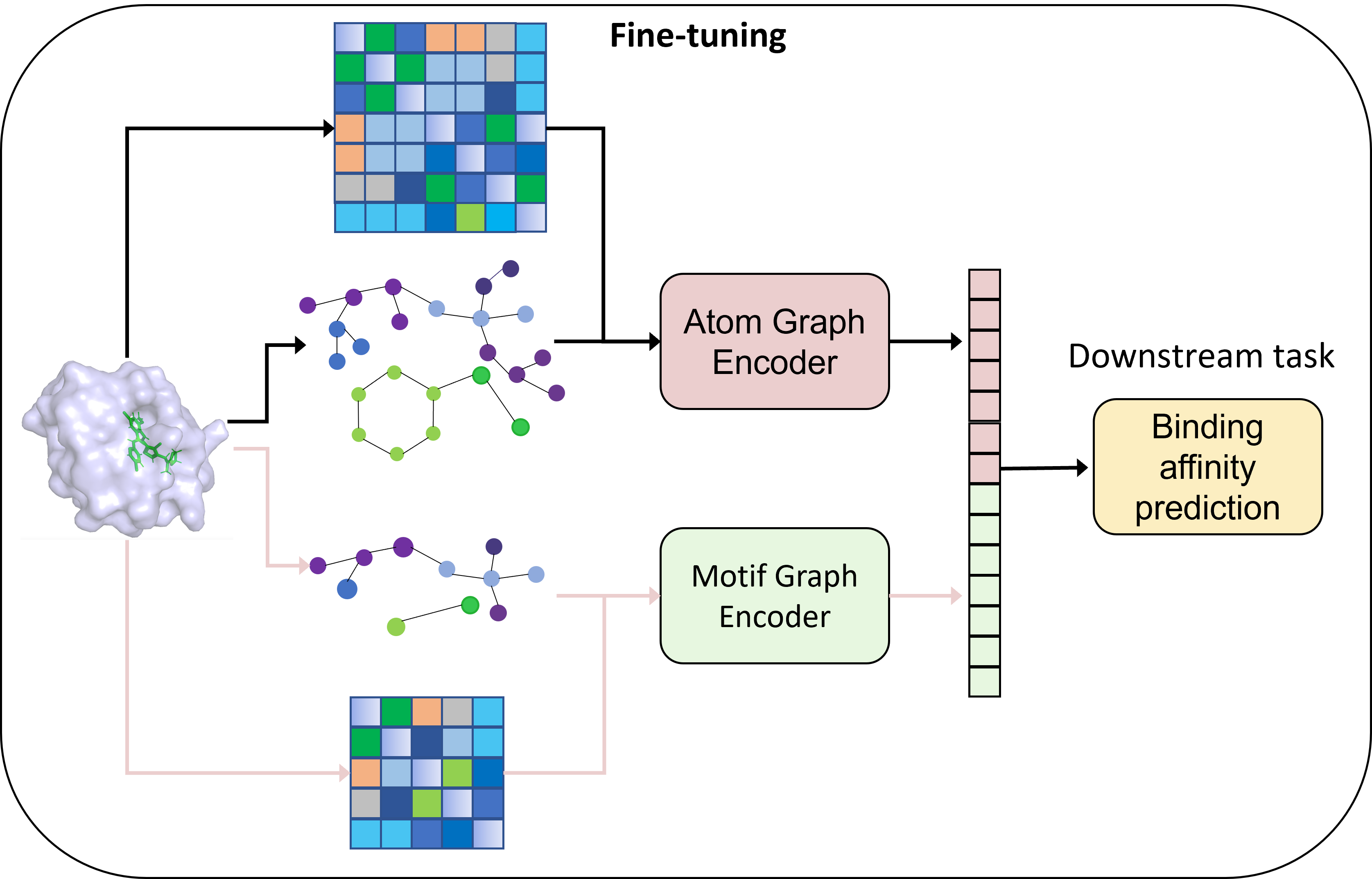}
\label{architecture-2}

}

\caption{(a) the pre-training process. In this phase, a complex structure is represented using atom graphs and motif graphs, while the distances between the nodes of the compound and protein are manually masked. (b) the fine-tuning process, during which complete information is provided for some downstream task. The output representations of the atom and motif graphs are then utilized to generate the final prediction.
} 

\label{architecture}
\end{figure*}

To address the above-mentioned problems, we propose a framework containing the pre-training model based on hierarchy graphs and the fine-tuning model, aiming to learn the interactions in different levels of biomolecules and achieve better performance in downstream tasks. 
Figure \ref{architecture} shows the framework of our model. The framework includes the pre-training part and fine-tuning part.

\subsection{Motif Graph}

\begin{figure}[htbp]
    \centering
    \includegraphics[width=0.95\linewidth]{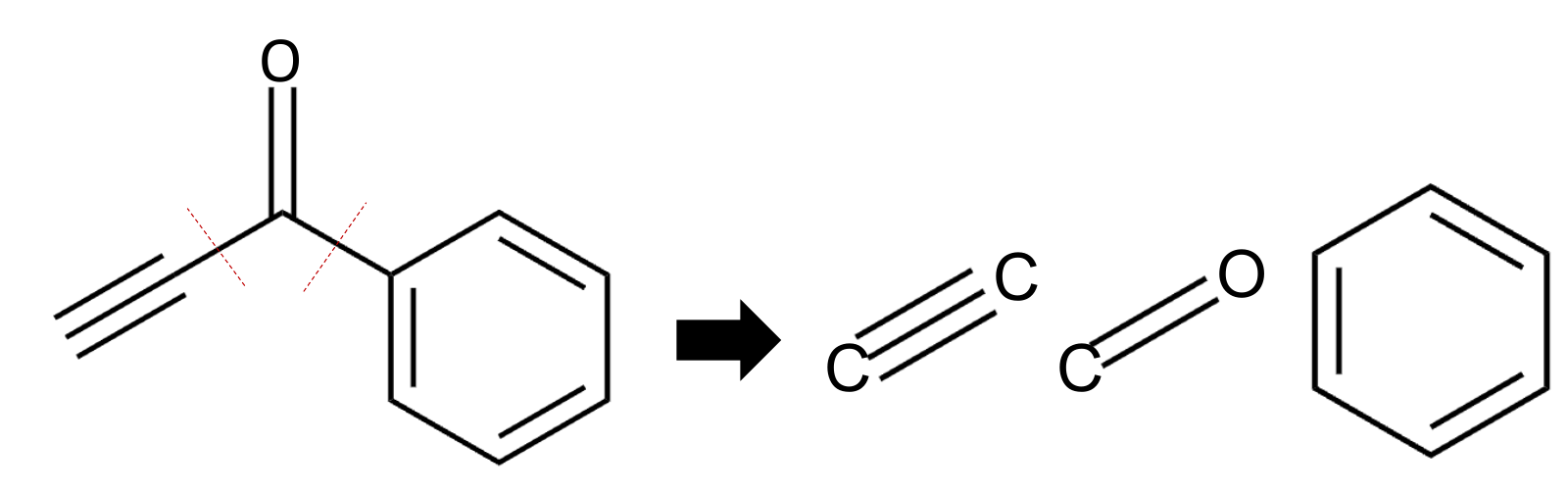}
    
    \vspace{8pt}
    
    \begin{tabular}{p{4cm}p{2cm}}
        \toprule
        \textbf{Bond Type} & \textbf{Action} \\
        \midrule
        single bond & break \\
        double bond \& triple bond & keep \\
        bond in a ring & keep \\
        \bottomrule
    \end{tabular}
    \caption{Motif graph generation rule}
    \label{fig:motif-gen}
\end{figure}

Given an atomic graph $G(V, E, P)$ where $V=\{v_1, v_2, \ldots, v_n\}$ represents atoms, $E=\{e_1, e_2, \ldots, e_n\}$ denotes bonds, and $P=\{p_1, p_2, \ldots, p_n\}$ signifies coordinates, we construct a motif graph $\mathrm{T}(M, E^{\prime}, Q)$ via function $\Theta(*)$:

\begin{equation}
\Theta(G): G(V, E, P) \to \mathrm{T}(M, E^{\prime}, Q)
\end{equation}

where $M=\{m_1, m_2, \ldots, m_n\}$, $Q=\{q_1, q_2, \ldots, q_n\}$, $q_i = \text{avg}\{p_1, p_2, \ldots, p_k\}$, $m_i=\{v_1, v_2, \ldots, v_k\}$, and $E^{\prime} \subseteq E$.

As shown in Figure \ref{fig:motif-gen}, we break single torsional bonds while retaining double bonds, triple bonds, and ring bonds. For proteins, we preserve the polypeptide backbone and only cleave sidechains, as compound interactions primarily occur at sidechains (Figure \ref{compare}A). Each motif's position is represented by its centroid, with initial embeddings averaged from constituent atoms.

\subsection{Graph Transformer} \label{sec:transformer}

\begin{figure*}[htbp] 
\centering
\begin{tcolorbox}[
    colback=white, 
    colframe=gray!80, 
    boxrule=1pt, 
    arc=24pt,              
    left=12pt,              
    right=12pt, 
    top=8pt,                
    bottom=8pt
]
    \centering
    \includegraphics[width=0.6\textwidth]{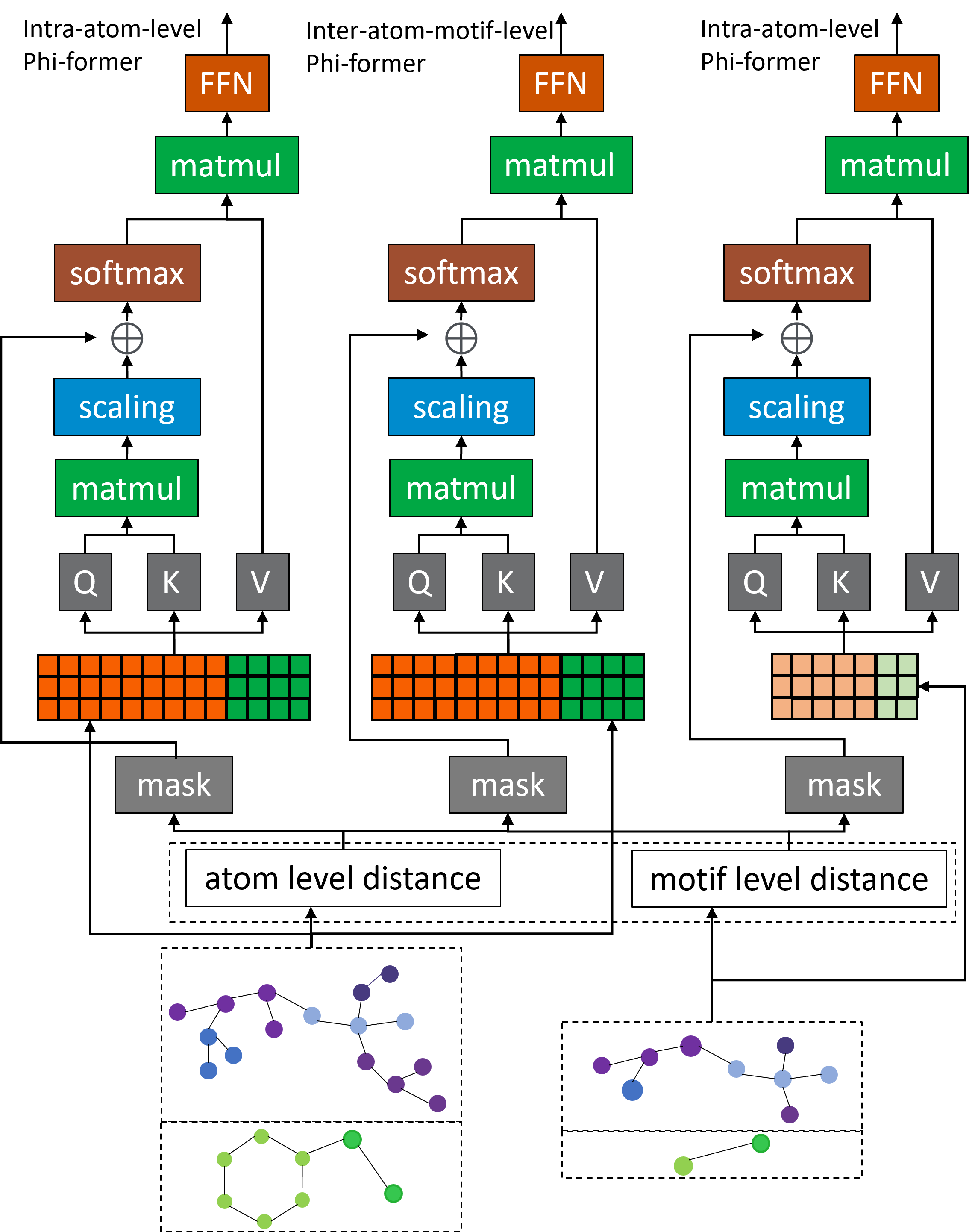}
\end{tcolorbox}
\caption{Graph transformer architecture as encoder} 
\label{fig:encoder}
\end{figure*}

We employ graph transformers \cite{b11} to encode both atomic and motif graphs into latent representations. Graph transformers are chosen over GNNs \cite{b7} for their ability to capture long-range interactions without over-smoothing and their larger parameter capacity suitable for pre-training. Figure \ref{fig:encoder} shows our encoder architecture.

\subsubsection{Spatial Positional Encoding (SPE)}
To distinguish nodes by their 3D coordinates while ensuring $E(n)$ invariance under rotation and translation, we encode Euclidean distances using Gaussian kernels. For distance $d_{ij} = Dis(v_{i}, v_{j})$:

\begin{equation}
    \boldsymbol{s}_{ij}^{k}=\{\mathcal{G}(\alpha t_{ij} d_{ij}+\beta t_{ij}, \mu^k, \sigma^k) \mid k \in[1, C]\}
\end{equation}

where $\mathcal{G}(d, \mu, \sigma) = \frac{1}{\sigma \sqrt{2 \pi}} e^{-\frac{(d-\mu)^2}{2 \sigma^2}}$, edge type $t_{ij}$ is determined by node types, and $\alpha$, $\beta$, $\mu^k$, $\sigma^k$ are trainable. We define mapping $\phi(G)$ that transforms input graph $G$ into embedding matrix $X$ and SPE $S$:

\begin{equation} \label{equation:phi}
    X, S = \phi(G)
\end{equation}

\subsubsection{Attention Bias}
We incorporate SPE as attention bias to distinguish spatial locations:

\begin{equation}
    attn(X)=\operatorname{softmax}\left(\frac{X W_q(X W_k)^{\top}}{\sqrt{d}}+S\right)
\end{equation}

For input $X$ with $m$ compound and $n$ protein nodes, $S$ contains compound-specific $S_c$ (upper-left), protein-specific $S_p$ (lower-right), and cross-distance $S_{cp}$, $S_{pc}$ (off-diagonal, populated during fine-tuning as detailed in Section \ref{sec:3-3}). Motif graphs use centroid positions with identical SPE computation.

With $X$ and $S$ defined, our graph transformer $GE$ produces node representations:

\begin{equation} \label{equation:ge}
   H = GE(X, S) = GE(\phi(G))
\end{equation}

where $H$ contains both compound and protein node representations.

\subsection{Pre-trainining} \label{sec:3-3}

We pre-train the model to understand hierarchical structural interactions in CPI tasks, incorporating unimolecular pre-training as necessary. We develop a distance-based self-supervised learning (SSL) task to elucidate the inter and intra relationships between various levels of components in CPI tasks. 

Given a complex conformation with spatial coordinates, we deliberately mask the intermolecular distance between the compound and the protein, effectively employing the protein as a reference. We apply random translations and rotations to the compound. We aim to model the docking process between the two rigid graphs at distinct hierarchical levels. Consequently, we introduce the SSL loss functions to facilitate this learning process.

Given a graph encoder, we can obtain the graph representation for every node, $H_{v^p}, H_{v^c} = GE(\phi(G))$, and $H_{m^p}, H_{m^l} = GE(\phi(\mathcal{T}))$. We define the predicted distance $S$ as follows:

\begin{equation}
    S\left(v^p, v^c\right)=f\left(h_{v^p}, h_{v^c}\right)
\end{equation}
where $v^p$ represents the node in the protein graph and $v^c$ denotes the node in the compound graph. The function $f(*)$ is the inference mechanism that maps the nodes' representations to the distance between the two corresponding nodes.

Our model aims to capture the interactions at both atomic and motif levels. To achieve this, we employ two separate encoders for these levels. However, since atoms and motifs are interrelated and cannot be considered in isolation. For example, the motif should be a restriction of the atom on the motif. Thus, we introduce a third encoder. This encoder treats the motif as prior knowledge for the atom and subsequently encodes the atom accordingly. During the pre-training stage, we incorporate three interaction losses associated with the three encoders: an atomic distance loss, a motif distance loss, and an atomic distance loss conditioned on motif distance. This approach allows our model to effectively learn the complex interplay between atoms and motifs and their contributions to the overall compound-protein interactions. We define the loss as

\begin{equation}
    L=L_V+L_M+L_{V \mid M}
\end{equation}

where 
\begin{equation}
L_V = \sum_{i=1}^m \sum_{j=1}^n (Dis(v^{c}_{i}, v^{p}_{j})- S(v^{c}_{i}, v^{p}_{j}))^2
\end{equation}

denotes the atomic level intra-loss. We mask the distance between compound nodes and protein nodes on the atom graph. The SSL task is to predict the masked distance.
\begin{equation}
L_M = \sum_{i=1}^m \sum_{j=1}^n (Dis(m^{c}_{i}, m^{p}_{j})- S(m^{c}_{i}, m^{p}_{j}))^2
\end{equation}

denotes the motif level intra-loss. We mask the distance between compound nodes and protein nodes on the motif graph. The SSL task is to predict the masked distance.
\begin{equation}
L_{V \mid M} = \sum_{i=1}^m \sum_{j=1}^n (Dis(v^{c}_{i}, v^{p}_{j})- S(v^{c}_{i}, v^{p}_{j}\mid S(M^c, M^p)))^2
\end{equation}

\begin{figure*}[htbp] 
\centering
\subfigure[$\pi$-$\pi$ interaction ground truth]{
\includegraphics[width=0.3\textwidth]{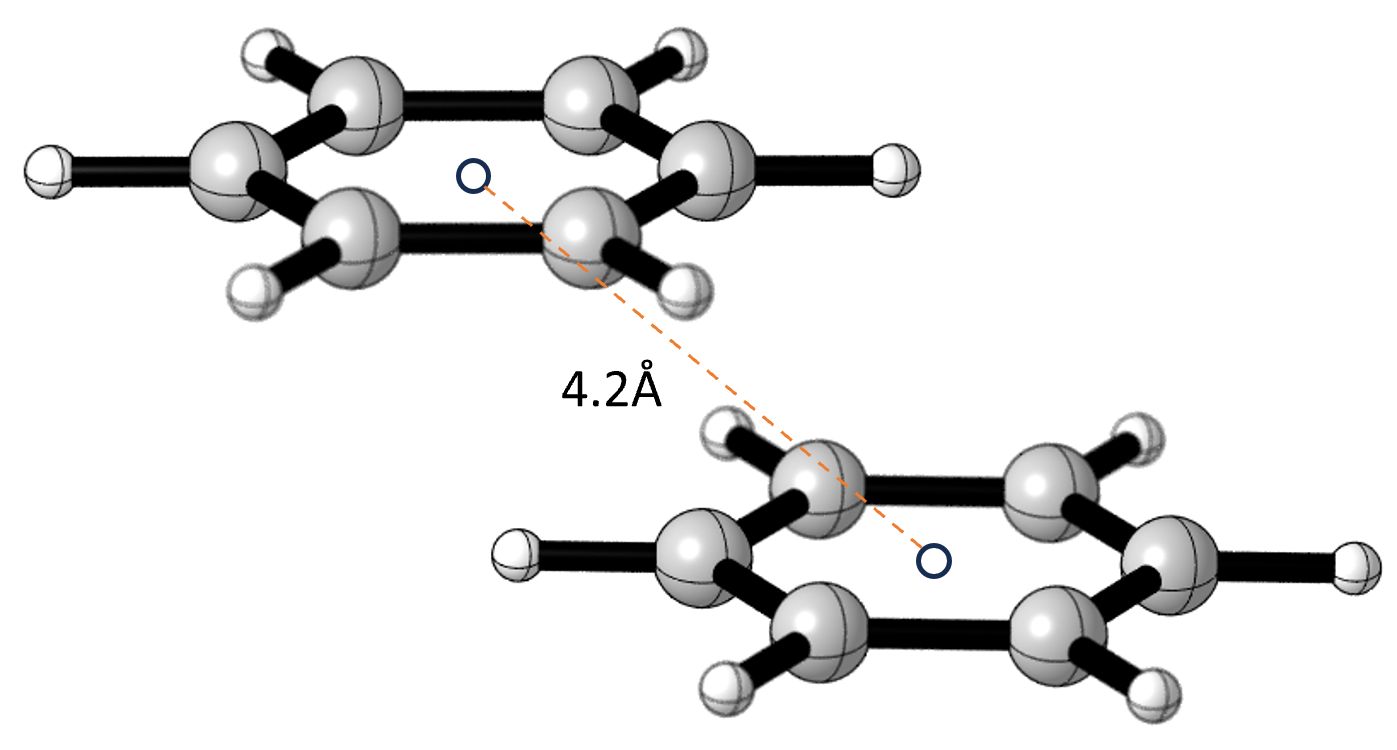}
\label{fig:case-1}
}
\hfill
\subfigure[results constraint by motif level model]{
\includegraphics[width=0.3\textwidth]{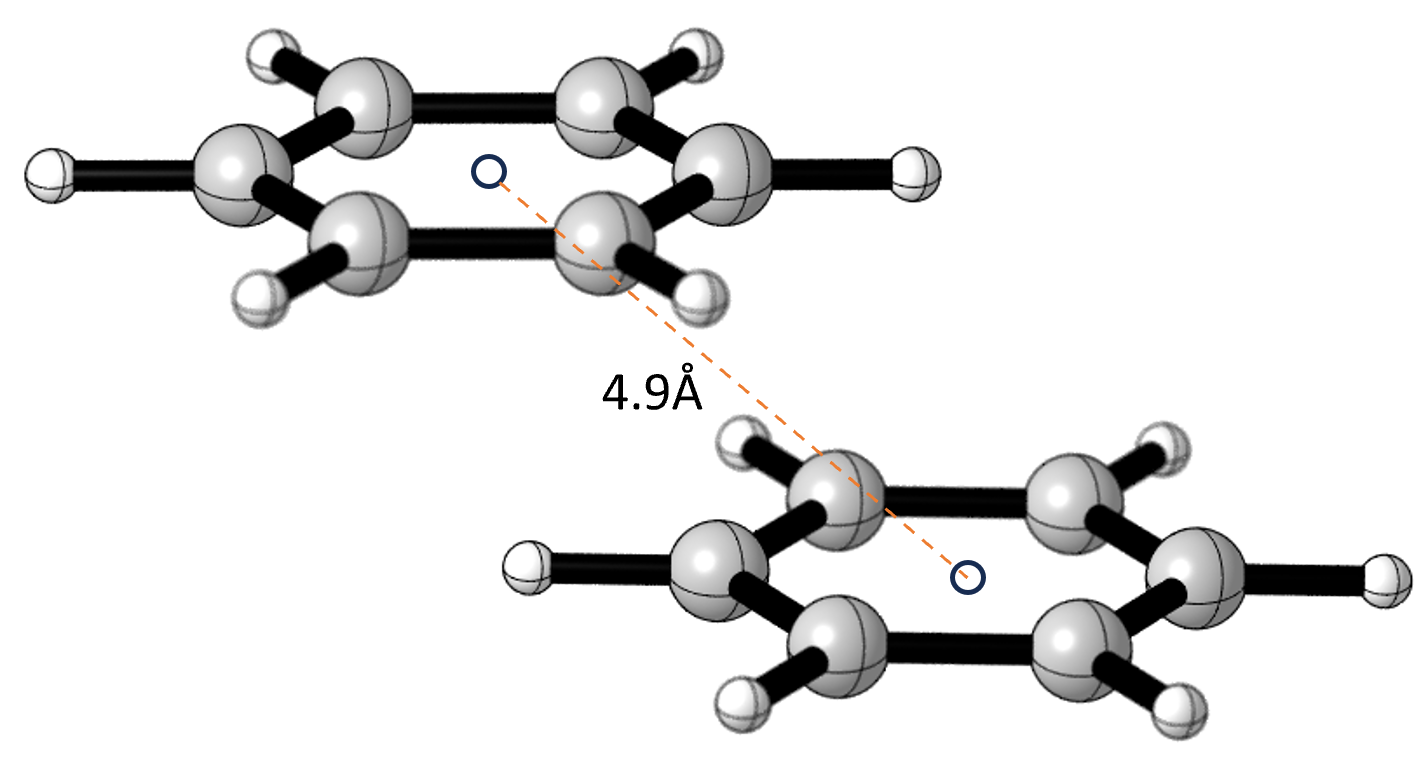}
\label{fig:case-2}
}
\hfill
\subfigure[results only on atomic level]{
\includegraphics[width=0.22\textwidth]{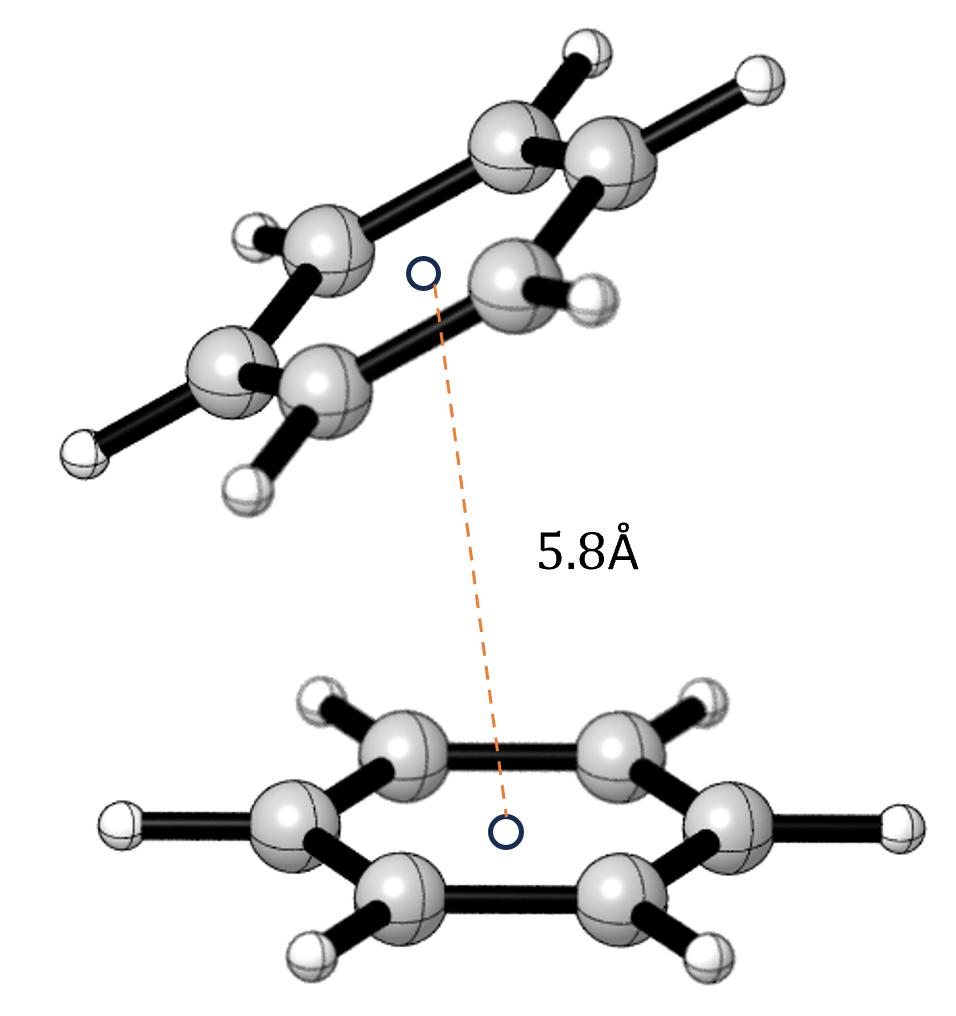}
\label{fig:case-3}
}
\caption{Case study showing motif constraints enable correct non-covalent interaction modeling.}
\label{fig:case}
\end{figure*}

represents the inter-loss between the atomic level and motif level. We mask the distance between compound nodes and protein nodes on the atom graph; however, we retain the motif distance as a priori knowledge, which are denoted as $M^{c}$ and $M^{p}$. establishing the relationship between different hierarchical levels. The SSL task is to predict the masked distance.

\subsection{Fine-tuning}
In the experimental section, we assess the performance of our proposed model on a classic CPI-related task, namely binding affinity prediction. Additionally, we conduct a case study to demonstrate that our model exhibits a higher degree of self-consistency with respect to established chemical rules.
\subsubsection{Binding Affinity Prediction}
Binding affinity is a measure that quantifies the strength of interaction between a compound and a protein. This task involves predicting the binding affinity as a continuous value, thus making it a regression task. After learning representations with the Encoder, we concatenate them on the three encoders to capture complex interactions between the compound and protein. We can easily predict binding affinity as a continuous value using a simple linear head. We defined the binding affinity loss by mean square error~(MSE).

%% file: sections/experiment.tex
\section{Experiment}

\subsection{Binding Affinity Prediction}
\subsubsection{Dataset}
We use PDBBind 2019 \cite{b30} for training and CASF-2016 \cite{b31} for testing. After removing CASF-2016 overlap and invalid data (protein-compound distance $>$ \qty{6}{\angstrom}), we obtain 16,493 training pairs (9:1 train-validation split) and 275 test pairs. We predict $pKa$ values to quantify binding affinity.
\begin{table}[htbp]
\centering
\caption{Results on Binding Affinity Prediction Task}
\renewcommand{\arraystretch}{1.3}
\rowcolors{2}{gray!15}{white}  
\begin{tabular}{lcc}
\toprule
\textbf{Method} & \textbf{RMSE} $(\downarrow)$ & \textbf{$R_p$} $(\uparrow)$ \\
\midrule
SIGN \cite{b28} & 1.316 & 0.797 \\
IGN \cite{b23} & 1.291 & 0.811 \\
$\triangle$VinaRF$_{20}$ \cite{b29} & -- & 0.816 \\
OnionNet \cite{b25} & 1.278 & 0.816 \\
Mol-PSI \cite{b26} & 1.278 & 0.844 \\
GraphDTA \cite{b24} & 1.562 & 0.697 \\
$K_{Deep}$ \cite{b27} & 1.270 & 0.820 \\
SS-GNN \cite{b5} & 1.181 & \textbf{0.853} \\
\midrule
\hiderowcolors  
\rowcolor{blue!8} Pre-train baseline & 1.418 & 0.758 \\
\rowcolor{blue!8} Phi-former (no pre-train) & 1.379 & 0.771 \\
\rowcolor{blue!15} \textbf{Phi-former (ours)} & \textbf{1.159} & 0.846 \\
\bottomrule
\end{tabular}
\label{table:affinity_results}
\end{table}
\subsubsection{Baselines and Metrics}
We compare against competitive drug-target binding affinity methods. To isolate our pre-training contribution, we establish two baselines: (1) a Uni-Mol-based pre-trained baseline \cite{b13} using similar prediction architecture, and (2) Phi-former trained from scratch with identical parameters. Evaluation metrics are Pearson correlation ($R_p$) and RMSE.

\subsubsection{Results and Analysis}
Table~\ref{table:affinity_results} shows Phi-former obtains the lowest RMSE (1.159) and a competitive Pearson correlation (0.846, second to SS-GNN's 0.853). Compared to the pre-trained baseline, our pairwise hierarchical pre-training significantly improves performance, demonstrating that modeling interactions during pre-training is crucial. The improvement over the non-pretrained version validates the effectiveness of our pre-training scheme on both metrics.

\subsection{Case Study: $\pi$-$\pi$ Interaction}
To verify our model's chemical consistency, we examine its ability to capture $\pi$-$\pi$ interactions—non-covalent forces between aromatic rings. Figure~\ref{fig:case-1} shows a classical $\pi$-$\pi$ interaction case. Figure~\ref{fig:case} reveals that the atom-only model (Figure~\ref{fig:case-3}) predicts \qty{6}{\angstrom} separation, failing to detect the interaction. In contrast, our motif-constrained model (Figure~\ref{fig:case-2}) correctly infers the spatial positioning of the aromatic rings, demonstrating chemically consistent predictions. This validates that hierarchical modeling enables our model to learn significant non-covalent interactions while adhering to established chemical principles.

%% file: IEEE-conference-template-062824.bbl
\begin{thebibliography}{00}

\bibitem{cpi1} B. X. Du, Y. Qin, Y. F. Jiang, Y. Xu, S. M. Yiu, H. Yu, and J. Y. Shi, ``Compound--protein interaction prediction by deep learning: databases, descriptors and models,'' \textit{Drug Discovery Today}, vol. 27, no. 5, pp. 1350--1366, 2022.

\bibitem{cpi2} S. Lim, Y. Lu, C. Y. Cho, I. Sung, J. Kim, Y. Kim, \textit{et al.}, ``A review on compound-protein interaction prediction methods: data, format, representation and model,'' \textit{Computational and Structural Biotechnology Journal}, vol. 19, pp. 1541--1556, 2021.

\bibitem{cpi3} H. Yabuuchi, S. Niijima, H. Takematsu, T. Ida, T. Hirokawa, T. Hara, \textit{et al.}, ``Analysis of multiple compound--protein interactions reveals novel bioactive molecules,'' \textit{Molecular Systems Biology}, vol. 7, no. 1, pp. 472, 2011.

\bibitem{cpi4} R. J. Brennan, T. Nikolskya, and S. Bureeva, ``Network and pathway analysis of compound--protein interactions,'' in \textit{Chemogenomics: Methods and Applications}, Totowa, NJ: Humana Press, 2009, pp. 225--247.

\bibitem{cpi5} W. Huber, ``A new strategy for improved secondary screening and lead optimization using high-resolution SPR characterization of compound--target interactions,'' \textit{Journal of Molecular Recognition: An Interdisciplinary Journal}, vol. 18, no. 4, pp. 273--281, 2005.



\bibitem{b2} W. L. Jorgensen and L. L. Thomas, ``Perspective on free-energy perturbation calculations for chemical equilibria,'' \textit{J. Chem. Theory Comput.}, vol. 4, no. 6, pp. 869--876, Jun. 2008.

\bibitem{b3} H. St\"{a}rk, O. Ganea, L. Pattanaik, R. Barzilay, and T. Jaakkola, ``Equibind: Geometric deep learning for drug binding structure prediction,'' in \textit{Proc. Int. Conf. Mach. Learn.}, 2022, pp. 20503--20521.

\bibitem{b4} W. Lu, Q. Wu, J. Zhang, J. Rao, C. Li, and S. Zheng, ``Tankbind: Trigonometry-aware neural networks for drug-protein binding structure prediction,'' \textit{bioRxiv}, 2022. [Online]. Available: https://doi.org/10.1101/2022.06.06.495043

\bibitem{b5} S. Zhang, Y. Jin, T. Liu, Q. Wang, Z. Zhang, S. Zhao, and B. Shan, ``SS-GNN: A simple-structured graph neural network for affinity prediction,'' \textit{ACS Omega}, vol. 8, no. 25, pp. 22496--22507, Jun. 2023.

\bibitem{b6} L. Huang, J. Lin, R. Liu, Z. Zheng, L. Meng, X. Chen, X. Li, and K.-C. Wong, ``CoaDTI: multi-modal co-attention based framework for drug--target interaction annotation,'' \textit{Brief. Bioinform.}, vol. 23, no. 6, pp. bbac446, Nov. 2022.

\bibitem{b7} T. N. Kipf and M. Welling, ``Semi-supervised classification with graph convolutional networks,'' \textit{arXiv preprint arXiv:1609.02907}, 2016.

\bibitem{b8} W. Hamilton, Z. Ying, and J. Leskovec, ``Inductive representation learning on large graphs,'' \textit{Adv. Neural Inf. Process. Syst.}, vol. 30, pp. 1024--1034, 2017.

\bibitem{b9} J. Gilmer, S. S. Schoenholz, P. F. Riley, O. Vinyals, and G. E. Dahl, ``Neural message passing for quantum chemistry,'' in \textit{Proc. Int. Conf. Mach. Learn.}, 2017, pp. 1263--1272.

\bibitem{b10} K. Xu, W. Hu, J. Leskovec, and S. Jegelka, ``How powerful are graph neural networks?'' \textit{arXiv preprint arXiv:1810.00826}, 2018.

\bibitem{b11} C. Ying, T. Cai, S. Luo, S. Zheng, G. Ke, D. He, Y. Shen, and T.-Y. Liu, ``Do transformers really perform badly for graph representation?'' \textit{Adv. Neural Inf. Process. Syst.}, vol. 34, pp. 28877--28888, 2021.

\bibitem{b12} S. Luo, T. Chen, Y. Xu, S. Zheng, T.-Y. Liu, L. Wang, and D. He, ``One transformer can understand both 2d \& 3d molecular data,'' \textit{arXiv preprint arXiv:2210.01765}, 2022.

\bibitem{b13} G. Zhou, Z. Gao, Q. Ding, H. Zheng, H. Xu, Z. Wei, L. Zhang, and G. Ke, ``Uni-mol: A universal 3d molecular representation learning framework,'' in \textit{Proc. Int. Conf. Learn. Represent.}, 2023.



\bibitem{b15} Z. Yu and H. Gao, ``Molecular representation learning via heterogeneous motif graph neural networks,'' in \textit{Proc. Int. Conf. Mach. Learn.}, 2022, pp. 25581--25594.

\bibitem{b16} X. Zang, X. Zhao, and B. Tang, ``Hierarchical molecular graph self-supervised learning for property prediction,'' \textit{Commun. Chem.}, vol. 6, no. 1, pp. 34, Mar. 2023.

\bibitem{b17} L. Dou, Z. Zhang, Y. Qian, Q. Zhang, et al., ``BCM-DTI: A fragment-oriented method for drug--target interaction prediction using deep learning,'' \textit{Comput. Biol. Chem.}, vol. 104, pp. 107844, Jun. 2023.

\bibitem{b18} D. Bui-Thi, E. Rivi\`{e}re, P. Meysman, and K. Laukens, ``Predicting compound-protein interaction using hierarchical graph convolutional networks,'' \textit{PLoS One}, vol. 17, no. 7, pp. e0258628, Jul. 2022.

\bibitem{b19} K. Bleakley and Y. Yamanishi, ``Supervised prediction of drug--target interactions using bipartite local models,'' \textit{Bioinformatics}, vol. 25, no. 18, pp. 2397--2403, Sep. 2009.

\bibitem{b20} F. Wan and J. Zeng, ``Deep learning with feature embedding for compound-protein interaction prediction,'' \textit{bioRxiv}, 2016. [Online]. Available: https://doi.org/10.1101/086033

\bibitem{b21} D. Weininger, ``SMILES, a chemical language and information system. 1. Introduction to methodology and encoding rules,'' \textit{J. Chem. Inf. Comput. Sci.}, vol. 28, no. 1, pp. 31--36, Feb. 1988.

\bibitem{b22} S. Li, F. Wan, H. Shu, T. Jiang, D. Zhao, and J. Zeng, ``MONN: a multi-objective neural network for predicting compound-protein interactions and affinities,'' \textit{Cell Syst.}, vol. 10, no. 4, pp. 308--322, Apr. 2020.

\bibitem{b23} D. Jiang, C.-Y. Hsieh, Z. Wu, Y. Kang, J. Wang, E. Wang, B. Liao, C. Shen, L. Xu, J. Wu, et al., ``Interactiongraphnet: A novel and efficient deep graph representation learning framework for accurate protein--ligand interaction predictions,'' \textit{J. Med. Chem.}, vol. 64, no. 24, pp. 18209--18232, Dec. 2021.

\bibitem{b24} T. Nguyen, H. Le, T. P. Quinn, T. Nguyen, T. D. Le, and S. Venkatesh, ``GraphDTA: predicting drug--target binding affinity with graph neural networks,'' \textit{Bioinformatics}, vol. 37, no. 8, pp. 1140--1147, Apr. 2021.

\bibitem{b25} L. Zheng, J. Fan, and Y. Mu, ``Onionnet: a multiple-layer intermolecular-contact-based convolutional neural network for protein--ligand binding affinity prediction,'' \textit{ACS Omega}, vol. 4, no. 14, pp. 15956--15965, Jul. 2019.

\bibitem{b26} P. Jiang, Y. Chi, X.-S. Li, Z. Meng, X. Liu, X.-S. Hua, and K. Xia, ``Molecular persistent spectral image (Mol-PSI) representation for machine learning models in drug design,'' \textit{Brief. Bioinform.}, vol. 23, no. 1, pp. bbab527, Jan. 2022.

\bibitem{b27} J. Jim\'{e}nez, M. Skalic, G. Martinez-Rosell, and G. De Fabritiis, ``K deep: protein--ligand absolute binding affinity prediction via 3d-convolutional neural networks,'' \textit{J. Chem. Inf. Model.}, vol. 58, no. 2, pp. 287--296, Feb. 2018.

\bibitem{b28} S. Li, J. Zhou, T. Xu, L. Huang, F. Wang, H. Xiong, W. Huang, D. Dou, and H. Xiong, ``Structure-aware interactive graph neural networks for the prediction of protein-ligand binding affinity,'' in \textit{Proc. 27th ACM SIGKDD Conf. Knowl. Discov. Data Min.}, 2021, pp. 975--985.

\bibitem{b29} F. Zhu, X. Zhang, J. E. Allen, D. Jones, and F. C. Lightstone, ``Binding affinity prediction by pairwise function based on neural network,'' \textit{J. Chem. Inf. Model.}, vol. 60, no. 6, pp. 2766--2772, Jun. 2020.

\bibitem{b30} R. Wang, X. Fang, Y. Lu, and S. Wang, ``The PDBbind database: Collection of binding affinities for protein-ligand complexes with known three-dimensional structures,'' \textit{J. Med. Chem.}, vol. 47, no. 12, pp. 2977--2980, Jun. 2004.

\bibitem{b31} M. Su, Q. Yang, Y. Du, G. Feng, Z. Liu, Y. Li, and R. Wang, ``Comparative assessment of scoring functions: the CASF-2016 update,'' \textit{J. Chem. Inf. Model.}, vol. 59, no. 2, pp. 895--913, Feb. 2019.

\end{thebibliography}
